\documentclass[10pt,twocolumn,letterpaper]{article}

\usepackage[pagenumbers]{cvpr} % To force page numbers, e.g. for an arXiv version

\usepackage{times}

\usepackage{url}
\usepackage{multirow}
\usepackage[T1]{fontenc}    % use 8-bit T1 fonts
\usepackage{url}            % simple URL typesetting
\usepackage{booktabs}       % professional-quality tables
\usepackage{amsfonts}       % blackboard math symbols
\usepackage{nicefrac}       % compact symbols for 1/2, etc.
\usepackage{microtype}      % microtypography
\usepackage{xcolor}         % colors
\usepackage{bbding}
\usepackage{amsmath}
\usepackage{amssymb}
\usepackage{mathtools}
\usepackage{amsthm}
\usepackage{bm}
\usepackage{tcolorbox}
\usepackage{xcolor} % 代码高亮显示 % 字体加颜色
\usepackage{color} % 字体加颜色
\usepackage{soul}
\usepackage{pifont}
\usepackage{xspace}
\usepackage[font={small}]{caption}
\usepackage{enumitem}
\usepackage{wrapfig,lipsum,booktabs}
\usepackage{float}
\usepackage[accsupp]{axessibility}  % Improves PDF readability for those with disabilities.

\definecolor{cvprblue}{rgb}{0.21,0.49,0.74}
\usepackage[pagebackref,breaklinks,colorlinks,allcolors=cvprblue]{hyperref}

\def\confName{CVPR}
\def\confYear{2026}

\title{MeshLAM: Feed-Forward One-Shot Animatable Textured Mesh Avatar Reconstruction}

\author{Yisheng He \quad Steven Hoi \\
Tongyi Lab, Alibaba Group\\
}

\begin{document}
\maketitle
\begin{abstract}
We introduce MeshLAM, a feed-forward framework for one-shot animatable mesh head reconstruction that generates high-fidelity, animatable 3D head avatars from a single image. Unlike previous work that relies on time-consuming test-time optimization or extensive multi-view data, our method produces complete mesh representations with inherent animatability from a single image in a single forward pass. Our approach employs a dual shape and texture map architecture that simultaneously processes mesh vertices and texture map with extracted image features from a shared transformer backbone, allowing for coherent shape carving and appearance modeling. To prevent mesh collapse and ensure topological integrity during feed-forward deformation, we propose an iterative GRU-based decoding mechanism with progressive geometry deformation and texture refinement, coupled with a novel reprojection-based texture guidance mechanism that anchors appearance learning to the input image. Extensive experiments demonstrate that our method outperforms state-of-the-art approaches in reconstruction quality, animation capability, and computational efficiency. Project page at \url{https://meshlam.github.io}.

\end{abstract}    
\section{Introduction}
\label{sec:intro}

Digital human avatars are central to applications in virtual reality, gaming, telepresence, and digital content creation. Despite significant advances in 3D head reconstruction, generating high-fidelity, directly animatable 3D avatars from a single image remains an open challenge. Existing methods face critical trade-offs: 2D-based approaches~\cite{SadTalker,StyleHEAT,Pirenderer,DBLP:conf/cvpr/WangM021} lack explicit 3D structure and suffer from multi-view consistency problem, while many 3D-aware models~\cite{DBLP:conf/cvpr/ZhangZLHLWGCL024,HAvatar} rely on expensive per-person optimization or require multi-view supervision, limiting their scalability and usability.

Recent work in 3D Gaussian Splatting~\cite{GaussianSplatting,hu2025forge4d} has demonstrated impressive rendering quality and real-time performance. However, most Gaussian-based avatar methods~\cite{GaussianAvatar} face significant practical limitations: they typically require per-subject video sequences for personalization. While recent feed-forward frameworks~\cite{LAM_Sig25,panolam} have introduced transformer-based architectures to generate animatable Gaussian heads using FLAME~\cite{FLAME:SiggraphAsia2017} priors, they suffer from two fundamental limitations. First, modeling fine-scale appearance details, such as hair strands, facial hair, or tattoos, demands an excessive number of Gaussian primitives, substantially increasing computational overhead during both training and inference. Second, optimizing such a large quantity of Gaussians within a single feed-forward pass proves challenging, often resulting in blurry reconstructions that lack high-frequency details. On the other hand, mesh-based representations offer a compelling alternative by naturally decomposing the reconstruction problem into geometric and appearance domains through vertices and texture maps. This decomposition provides inherent advantages for optimization and computational efficiency. However, existing mesh-based 3DMM reconstruction methods~\cite{DBLP:conf/eccv/KhakhulinSLZ22,DBLP:conf/cvpr/XuYCWDJT20} fail to recover detailed geometry beyond the facial region, particularly for hair and headwear, and typically do not reconstruct high-fidelity textures, limiting their practical utility for creating complete animatable digital avatars.

In this work, we introduce a novel feed-forward framework for one-shot animatable mesh head reconstruction that generates complete textured mesh avatars in seconds through a single forward pass. Our approach synergistically combines the structural advantages of mesh representations with the representational power of transformers, achieving high-fidelity reconstruction without requiring iterative test-time optimization. At the core of our method lies a dual-branch architecture that explicitly decouples shape and appearance learning: one branch predicts per-vertex deformations relative to a FLAME template to capture geometric details, while the other synthesizes a high-resolution UV-aligned texture map for photorealistic surface appearance. This decomposition enables independent yet coherent modeling of geometry and appearance, facilitating detailed texture synthesis without overburdening the shape decoder with appearance details.
However, we observe that na\"ively training such a dual-path system end-to-end frequently leads to mesh degeneracies, particularly in regions requiring large deformation, such as long hair or headwearing. The fundamental challenge stems from the instability of direct vertex deformation offset regression. When vertices are displaced independently without structural constraints, errors can propagate into severe mesh distortions or topological collapse. To address this critical limitation, we introduce an iterative GRU-based decoding mechanism that progressively refines both mesh geometry and texture features through multiple coherent update steps. Each iteration leverages hidden states from previous steps, enabling stable coarse-to-fine reconstruction dynamics within a fully differentiable, learnable framework.

Crucially, to ground texture synthesis in observable appearance, we develop a reprojection module that dynamically warps the input image and the prediction error onto the evolving mesh surface at each iteration. By back-projecting visible pixels into UV space using the current geometry estimate, this strategy provides direct visual supervision for texture prediction while guiding shape deformation toward photometrically consistent solutions. The resulting closed-loop feedback between 3D geometry and 2D observation significantly enhances reconstruction realism and robustness, ensuring faithful detail reproduction while maintaining mesh integrity.

Our method thus departs from both optimization-heavy pipelines and fragile single-pass regressors, offering a balanced solution that is fast, stable, and expressive. In summary, our contributions are:
\begin{itemize}[leftmargin=*]
    \item We present a feed-forward network for one-shot animatable mesh head reconstruction, leveraging a dual-branch design that separately models vertex deformations and UV texture maps, enabling high-fidelity shape carving and appearance synthesis.
    
    \item We introduce an iterative GRU-based refinement mechanism that progressively deforms the mesh and refines texture, effectively alleviating mesh collapse and maintaining topological coherence.
    
    \item We propose a reprojection-based texture guidance strategy that unwraps the input image and the prediction error onto the deformed mesh at each step, providing strong visual cues for realistic texture generation and improving cross-domain generalization.
    
\end{itemize}
\vspace{-2mm}
\section{Related Work}
\label{related}

\subsection{2D-based Avatar Generation}
2D-based methods leverage powerful convolutional neural networks (CNNs)~\citep{DBLP:conf/cvpr/IsolaZZE17,DBLP:conf/nips/GoodfellowPMXWOCB14} and generative adversarial networks (GANs)~\citep{DBLP:conf/cvpr/StyleGAN} for direct image synthesis. Early approaches~\citep{DBLP:conf/cvpr/WangDYSW23,DBLP:conf/cvpr/BurkovPGL20,DBLP:conf/iccv/ZakharovSBL19} focused on injecting expression and pose features into generator networks, typically employing U-Net or StyleGAN architectures. Alternative 2D paradigms~\citep{DBLP:journals/corr/abs-2407-03168,DBLP:conf/cvpr/ZhangQZZW0CW023,DBLP:conf/cvpr/HongZS022,DBLP:conf/mm/DrobyshevCKILZ22,DBLP:conf/nips/SiarohinLT0S19} represent facial motions as warping fields applied to source images. 
The recent advent of diffusion models has significantly advanced 2D avatar quality~\citep{DBLP:journals/corr/abs-2410-07718,DBLP:journals/corr/abs-2406-08801,DBLP:conf/eccv/TianWZB24}, though these methods still face challenges in generation speed and computational efficiency. Audio-driven 2D approaches~\citep{DBLP:conf/cvpr/ZhangCWZSGSW23,DBLP:journals/corr/abs-2211-12368,DBLP:conf/iccv/GuoCLLBZ21} offer intuitive control but lack explicit manipulation of facial expressions and poses. 
A fundamental limitation of 2D methods is their inability to handle large pose or expression variations due to the absence of explicit 3D structure, often resulting in unrealistic distortions or identity inconsistencies. While some 2D techniques~\citep{SadTalker,StyleHEAT,Pirenderer,DBLP:conf/cvpr/WangM021,MegaPortraits} incorporate 3D Morphable Models (3DMMs)~\citep{DBLP:conf/fgr/GerigMBELSV18,DBLP:journals/tog/LiBBL017,DBLP:conf/avss/PaysanKARV09,DBLP:conf/siggraph/BlanzV99} to mitigate these issues, they typically cannot achieve true free-viewpoint rendering, limiting their applicability in immersive applications.
\vspace{-1mm}

\subsection{3D-based Avatar Generation}
3D-aware methods provide superior geometric consistency and enable free-viewpoint rendering capabilities. Mesh-based 3D approaches~\citep{DBLP:conf/eccv/KhakhulinSLZ22,DBLP:conf/cvpr/XuYCWDJT20,rome2022,liao2025SOAP} primarily relied on 3DMMs for head avatar reconstruction, while confronting missing head regions like hair or requiring time-consuming per-person optimization. The emergence of Neural Radiance Fields (NeRFs)~\citep{DBLP:conf/eccv/MildenhallSTBRN20} catalyzed a shift toward higher-fidelity representations, with numerous recent methods~\citep{DBLP:conf/siggraph/YuFZWYBCSWSW23,DBLP:conf/cvpr/MaZQLZ23,DBLP:conf/cvpr/LiZWZ0CZWB023,DBLP:conf/eccv/KiMC24,DBLP:conf/cvpr/BaiFWZSYS23,PointAvatar,Nerfies,INSTA} adopting this paradigm for capturing fine details such as hair.
However, NeRF-based approaches~\citep{DBLP:conf/cvpr/ZhangZLHLWGCL024,HAvatar,DBLP:conf/cvpr/BaiTHSTQMDDOPTB23,AD-NeRF,DBLP:journals/tog/GaoZXHGZ22,DBLP:journals/tog/ParkSHBBGMS21,DBLP:conf/cvpr/AtharXSSS22,DBLP:journals/corr/abs-2112-05637,DBLP:conf/iccv/TretschkTGZLT21,DBLP:conf/cvpr/GafniTZN21} often require multi-view imagery or single-view videos, limiting generalization to unseen identities. Alternative strategies~\citep{DBLP:conf/cvpr/SunWWLZZL23,DBLP:conf/3dim/ZhuangMKS22,DBLP:journals/pami/SunWZHWL24,DBLP:journals/tvcg/TangZYZCMW24,DBLP:conf/iclr/XuZLZBFS23} circumvent this data requirement by training generators on random noise and subsequently inverting them for identity-specific reconstruction, while still being computation-consuming. The recent emergence of 3D Gaussian Splatting~\citep{GaussianSplatting} has introduced a promising alternative, combining high-quality rendering with real-time performance. However, most existing methods~\citep{GaussianAvatar,DBLP:conf/cvpr/XuCL00ZL24} typically require per-identity video data for training, limiting their generalization capability. 
Recently, feed-forward animatable Gaussian avatar generation frameworks~\cite{LAM_Sig25,GAGAvatar,panolam,avat3r} are introduced into this field. Specifically, LAM~\citep{LAM_Sig25} and its follow-up~\cite{panolam} introduce a feed-forward transformer to decode animatable Gaussian heads based on FLAME model. However, a large number of Gaussian points are required to model detail textures, which increases the computation costs in the transformer backbone. And it is hard to optimize detailed high-fidelity textures in a single forward pass, leading to blurry results. 
In contrast, our work generates animatable textured mesh avatars through a dual-branch architecture that separately optimizes vertex deformation and texture map reconstruction with iterative GRU decoders. This decomposition enables efficient shape modeling with a sparse set of vertices while preserving high-frequency appearance in a compact texture map.

\begin{figure*}[ht]
    \centering
    %\vspace{3in}
    \includegraphics[width=1.\linewidth]{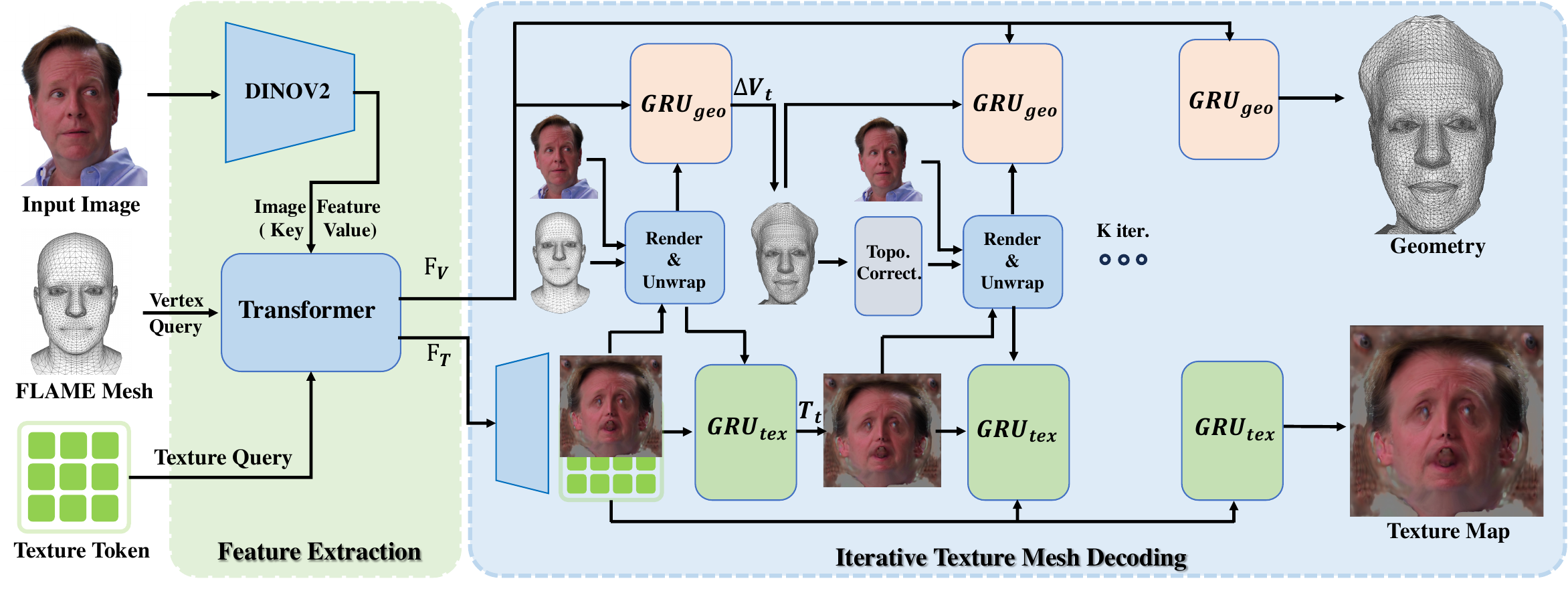}
    \caption{\textbf{Overall Framework.} Our method reconstructs an animatable 3D texture head mesh from a single image through dual shape and texture branches. After extracting features from the input image with a shared transformer, the shape branch predicts vertex deformations while the texture branch synthesizes UV texture maps, both conditioned on a FLAME template. Both branches are refined iteratively via GRU decoders with topology correction for shape refinement and reprojection guidance for texture enhancement.}
    \label{fig:framework}
    \vspace{-4mm}
\end{figure*}

\section{Methodology}
\label{sec:method}

\subsection{Overview}
\label{sec:overview}

Given a single input image, our framework reconstructs an animatable 3D head mesh in a feed-forward manner through a dual-branch transformer architecture. As shown in Fig.~\ref{fig:framework}, the network simultaneously predicts vertex deformations and a UV texture map, both conditioned on a FLAME template~\cite{FLAME:SiggraphAsia2017}. 
The shape branch processes initial mesh vertices via cross-attention to image features extracted by a ViT backbone~\cite{DinoV2}. The texture branch operates on a learnable token grid aligned with the FLAME UV map, which attends to the same image features using shared attention blocks. Both branches are refined over $T$ recurrent steps by GRU-based decoders that update hidden states conditioned on fused visual context.
At each step, a topology correction module is applied to preserve mesh regularity, while a reprojection mechanism samples the input image onto the current mesh estimate and unwarps it into UV space to guide texture refinement. After the final iteration, the output vertices form a detailed, animatable mesh, and the texture map provides a high-fidelity appearance.

\subsection{Dual Vertex and Texture Feature Extraction}
Given a single input image, our goal is to reconstruct an animatable, textured 3D head mesh by jointly predicting vertex deformations and a high-resolution texture map. We build upon the FLAME model as a parametric prior for identity, expression, and topology, providing a structured mesh template that supports standard rigging and animation. However, FLAME alone cannot represent fine geometric details such as hair, accessories, or facial textures. To overcome these limitations, we introduce a dual-branch framework that deforms the FLAME mesh to capture full-head geometry and synthesizes a UV-aligned texture map for appearance fidelity. 

\noindent\textbf{Image Feature Extraction.} We begin by extracting rich multi-scale image features $F_I \in \mathbb{R}^{N \times C}$ using a pre-trained DINOv2 backbone~\cite{DinoV2}. Following~\cite{DPTICCV2021,LAM_Sig25}, we aggregate feature maps from the 5th, 12th, 18th, and 24th transformer blocks, and fuse them via a lightweight MLP. The resulting fused feature map captures both local texture cues and global semantic structure.

\noindent\textbf{Vertex Feature Extraction.} The shape branch initializes with the FLAME mesh vertices $V_0$, augmented with positional encoding~\cite{NeRF} and passed through shared MLPs to form initial vertex features $F_V$. These features then interact with the image features $F_I$ through $L_\mathcal{A}$ cross-attention layers:
\begin{equation}
    F_{V_i} = \mathcal{A}_i(F_{V_{i-1}}, F_I),
    \label{eq:vertex_attn}
\end{equation}
where each layer enables vertices to attend to relevant image regions, progressively refining their contextual embeddings. After $L_\mathcal{A}$ layers, the final $F_V$ encodes geometry-aware features conditioned on the input.

\noindent\textbf{Texture Feature Extraction.} In parallel, the texture branch operates on a learnable token grid $T_0 \in \mathbb{R}^{H_t \times W_t \times C_t}$, aligned with the FLAME UV atlas. This grid is flattened into a sequence and processed through the same stack of cross-attention layers:
\begin{equation}
    F_{T_i} = \mathcal{A}_i(F_{T_{i-1}}, F_I),
    \label{eq:texture_attn}
\end{equation}
where $F_{T_{i}}$ represents the refined texture features at the output of the $i$-th cross-attention layer. Through this shared attention mechanism, the model learns to transfer rich, spatially-aware texture information from the image features $F_I$ to the final texture feature $F_T$, ultimately enabling the reconstruction of photorealistic surface appearance.

\subsection{Iterative Texture Mesh Decoding}
The extracted vertex and texture features now encapsulate rich geometric and semantic information necessary for high-quality shape and texture reconstruction. A straightforward approach would employ a shape decoder to predict vertex deformation offsets from the vertex features, thereby adapting the FLAME mesh to the target geometry, alongside a texture decoder to generate high-fidelity texture maps from the texture features.
However, we observe that this direct decoding paradigm frequently leads to mesh collapse, particularly in scenarios requiring large deformation, such as long hairstyles. The fundamental challenge lies in maintaining mesh topology and local geometric coherence when individual vertices undergo large, unconstrained displacements. 

To address this critical limitation, we introduce an iterative decoding mechanism that employs recurrent GRU-based update operators. This design enables progressive, coarse-to-fine deformation of the mesh geometry and simultaneous refinement of the texture map, ensuring stable optimization while preserving structural integrity throughout the reconstruction process. Given the extracted vertex features $F_V$ and texture features $F_T$, we now detail our iterative decoding framework for simultaneous mesh deformation and texture synthesis. 

\noindent\textbf{Texture Space Projection and Unwrapping.}
To provide strong visual guidance for texture synthesis and visual guidance for geometry deformation, we leverage the input image through a texture space projection mechanism. At each iteration $t$, we animate the current deformed mesh $M_t$ from canonical neutral space to match the expression parameters of the input image. The animated mesh is then rasterized to establish correspondence between image pixels and UV coordinates, enabling us to unwrap the input image and prediction error features into texture space:

\begin{equation}
\begin{aligned}
U_t = \mathcal{U}(I_{\text{input}}, \mathcal{R}(M_t^{\text{animated}})), 
\label{eq:texture_unwrap}
\end{aligned}
\end{equation}

where $\mathcal{R}$ denotes the rasterization operation and $\mathcal{U}$ the unwrapping operation, and $U_t$ represents the unwrapped texture map from the input image at iteration $t$. The rasterization and unwrapping operation $\mathcal{R}$ can also be utilized to unwrap the prediction error features $F_{d_t}$, This progressively updated unwrapping provides direct visual evidence of surface appearance, which is particularly crucial for visible regions where the input image offers reliable texture information.

\noindent\textbf{Iterative Texture Synthesis with Visual Guidance.}
Building upon the unwrapped texture $U_t$, we decode the texture features $F_T$ into high-resolution texture maps through an iterative refinement process that incorporates direct visual evidence. The process begins with a DPT decoder~\cite{DPTICCV2021} that upsamples and decodes the texture features $F_T$ to generate an initial texture map $T_0 \in \mathbb{R}^{H_a \times W_a \times 3}$ and a corresponding feature map $F_a \in \mathbb{R}^{H_a \times W_a \times C_t}$. This initialization provides a robust starting point for subsequent refinement. The following iterative texture decoder employs a convolutional GRU architecture, $\text{GRU}_\text{tex}$, that progressively enhances texture details through multi-source information integration at each iteration $t$:
\begin{equation}
\begin{aligned}
&T_{t+1} = \text{GRU}_\text{tex}\left(\varphi\left(\left[\varphi\left([T_t, U_t]\right), F_a, F_{d_t}\right]\right), h_{t}^\text{tex}\right), \\
&F_{d_t} = \mathcal{U}(\varphi([I_{\text{input}}, I_{\text{rendered}}, I_{\text{input}}-I_{\text{rendered}}]))
\end{aligned}
\label{eq:texture_refine}
\end{equation}
where $\varphi$ denotes convolutional operations, $[\cdot,\cdot]$ indicates feature concatenation, $T_t$ represents the previous texture estimate, $F_a$ the latent texture features, and $U_t$ the current unwrapped image texture. Another design is the prediction error feature $F_{d_t}$, computed by applying convolutional processing to the concatenation of the input image $I_\text{input}$ and the rendered image $I_\text{rendered}$ of the animated mesh textured with $T_t$, which are then unwrapped to the UV space with the unwrapping operation $\mathcal{U}$. This error signal captures the discrepancy between observed and synthesized appearance, providing direct feedback for texture correction.
This architecture enables the model to progressively fuse learned priors with observed appearance data, effectively balancing reconstruction fidelity in visible regions with plausible completion in occluded areas. The recurrent nature of the GRU allows the network to maintain temporal coherence across iterations while refining high-frequency details.

\noindent\textbf{Iterative Geometric Deformation.}
To achieve stable and topology-preserving mesh deformation, we employ a recurrent decoding mechanism that progressively refines vertex positions through a sequence of controlled updates. The process begins with initial deformation fields $\Delta V_{0}$ set to zero for all vertices, ensuring gradual deformation from the FLAME template.
At each iteration $t$, our geometry GRU, $\text{GRU}_\text{geo}$, processes multiple information sources to predict deformation offsets:
\begin{equation}
\Delta V_{t+1} = \text{GRU}_\text{geo}([\psi(\vartheta(V_t), F_{{d_t}2v}), F_V], h_{t}^\text{geo}),
\label{eq:geo_deform}
\end{equation}
where $\psi$ denotes MLP operations, $\vartheta$ represents positional encoding, $F_{{d_t}2v}$ contains visual prediction error features, $F_{{d_t}}$ projected to vertex space via UV coordinates, and $h_{t}^\text{geo}$ maintains deformation history. The vertex positions are then updated as $V_{t+1} = V_t + \Delta V_{t+1}$.

\textbf{Part-Aware Deformation.} To preserve crucial facial animation capabilities while allowing sufficient deformation for non-rigid components, we introduce semantic-aware deformation constraints. Specifically, we apply region-dependent clipping to $\Delta V_t$: larger deformation ranges, $\delta_\text{hair}$, are permitted for hair regions, moderate ranges, $\delta_\text{neck}$ and $\delta_\text{face}$, for neck and facial areas. The eyeball and eyelid vertices undergo non deformation to maintain anatomical correctness.

\textbf{Topology Correction.} Following each deformation step, we apply topology correction~\cite{continuous_remeshing,liao2025SOAP} to address potential mesh degradation. The correction pipeline comprises three operations: (1) subdividing triangles with edge lengths exceeding threshold $\varepsilon$, (2) flipping inconsistent triangle orientations, and (3) removing geometrically invalid faces. As remeshing alters vertex connectivity, we update skinning weights $W \in \mathbb{R}^{N\times|J|}$ and blendshapes $B \in \mathbb{R}^{500\times|J|}$ through barycentric interpolation from neighboring vertices. To preserve joint positions under changing topology, we recompute the joint regressor matrix as:
$J = J(M + B_s(\beta))^{-1},$
where $J$ represents canonical joints and $M$ denotes the updated template mesh. This ensures consistent skeletal animation despite topological changes.

This comprehensive approach enables complex non-rigid deformations while maintaining mesh integrity, anatomical correctness, and animation compatibility throughout the iterative refinement process.

\noindent\textbf{Coherent Multi-scale Refinement.}
The geometric deformation and texture synthesis processes operate in lockstep, with the mesh evolution directly influencing the texture unwrapping at each iteration. This creates a synergistic relationship where improved geometry leads to more accurate texture projection, which in turn provides better guidance for subsequent texture refinement. After $K$ iterations, we obtain the final deformed mesh $V_K$ and high-fidelity texture map $T_K$, which together form our complete 3D reconstruction. This iterative paradigm demonstrates superior stability compared to single-pass decoding, particularly for challenging cases requiring significant deformation from the template geometry, while the incorporation of progressive texture unwrapping ensures appearance consistency with the input image.

\noindent\textbf{Optional Neural Renderer.} To enhance the rendering results, our framework also support utilizing a neural network as an optional renderer. Given the final deformed textured mesh, we animate and rasterize it, getting the rendered RGB image $I_\text{rendered}$ together with the UV coordinate of each pixel, which then sample features from the texture feature map $F_a$ to the image plane as $F_\text{img}$, which are then concatenated with $I_\text{rendered}$ and pass into a StyleGan-like~\cite{GAGAvatar} UNet renderer for final rendered results.

\subsection{Optimization and Regularization}
\label{sec:loss}

To train our model end-to-end, for each iteration, we employ a multi-component objective that jointly optimizes geometry, texture, and semantic coherence through a combination of photometric, geometric, and structural losses. These terms are designed to ensure high-fidelity reconstruction, prevent mesh degeneration, and maintain anatomical plausibility without relying on external optimization at test time.

Our primary supervision comes from image-space reconstruction quality. We render the deformed and animated mesh $V_K$ with predicted texture $T_K$ and minimize both pixel-wise and perceptual discrepancies against ground truth images:
\begin{equation}
    \mathcal{L}_\text{img} = \|I_\text{rendered} - I_\text{gt}\|_2^2 + \phi(I_\text{rendered}, I_\text{gt}),
\end{equation}
where $\phi$ denotes the perceptual loss. This combined loss ensures sharp textures and global appearance consistency.

To improve foreground-background separation and silhouette accuracy, we supervise the rendered silhouette with the ground truth foreground mask:
\begin{equation}
    \mathcal{L}_\text{mask} = \|M_\text{rendered} - M_\text{gt}\|_2^2.
\end{equation}

We further enhance geometric accuracy by leveraging surface normals estimated from real images via a pretrained normal prediction network~\cite{sapiens}. We render the normal map from our reconstructed mesh and compare it to the pseudo-ground truth:
\begin{equation}
    \mathcal{L}_\text{normal} = \|N_\text{rendered} - N_\text{gt}\|_2^2,
\end{equation}
which encourages faithful recovery of fine-scale surface curvature.

To guide part-aware deformation and preserve anatomical structure, we introduce a semantic part loss. Using a pretrained face parsing model~\cite{dinu2022faceparsing}, we generate part segmentations (e.g., hair, eyes, mouth) for the input image and compare them with the projected segmentation of the rendered mesh. Specifically, we compute an MSE loss:
\begin{equation}
    \mathcal{L}_\text{part} = \|P_\text{rendered} - P_\text{gt}\|^2_2.
\end{equation}
This encourages correct alignment of different head components and prevents implausible deformations, such as distorted eye placement or misplaced hair boundaries.

Finally, to maintain mesh smoothness and prevent vertex scattering or self-intersections, we apply Laplacian regularization on the final mesh:
\begin{equation}
    \mathcal{L}_\text{lap} = \sum_{v_i} \left\| v_i - \frac{1}{|\mathcal{N}(i)|} \sum_{j \in \mathcal{N}(i)} v_j \right\|^2,
\end{equation}
where $\mathcal{N}(i)$ denotes the one-ring neighborhood of vertex $v_i$. This term promotes local geometric regularity and stabilizes training, particularly for regions undergoing large deformations.

The loss at the $t_\text{th}$ iteration is a weighted sum:
\begin{equation}
    \mathcal{L}_\text{t} = \lambda_{i} \mathcal{L}_\text{img} + \lambda_m \mathcal{L}_\text{mask} + \lambda_n \mathcal{L}_\text{normal} + \lambda_p \mathcal{L}_\text{part} + \lambda_l \mathcal{L}_\text{lap},
\end{equation}
with weights set empirically as $\lambda_i=1, \lambda_m=1$, $\lambda_n=1$, $\lambda_p=0.5$, $\lambda_l=2$. The total loss with $N$ iterations is then defined as: $\mathcal{L}_\text{total} = \sum_{t=1}^{N} {\gamma^{N-t}\mathcal{L}_t}$, where $N=2$ and $\gamma=0.8$ are set empirically in our experiments.

\begin{figure*}[ht]
    \centering
    \includegraphics[width=\linewidth]{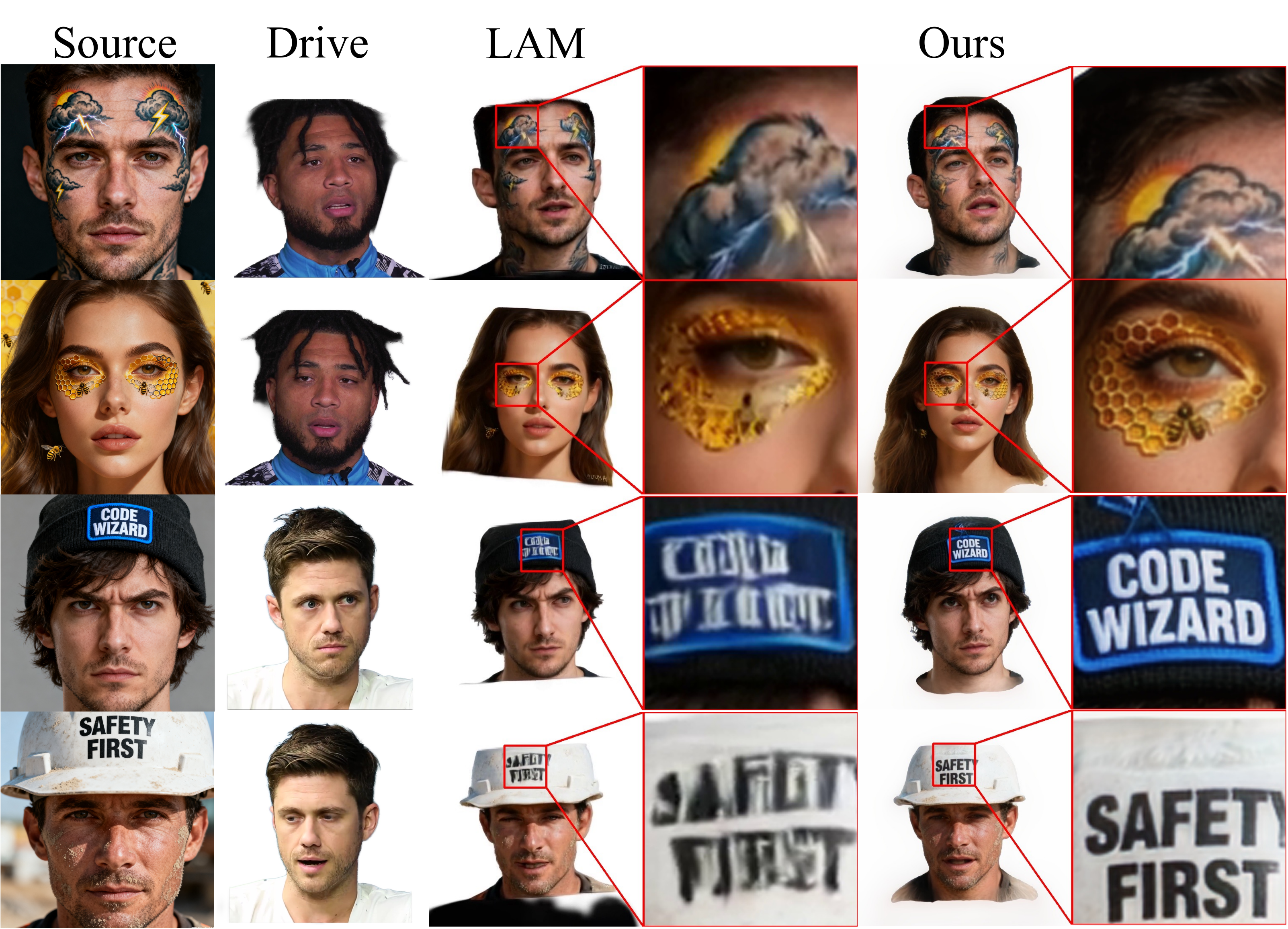}
    \caption{Qualitative comparison of 3D head avatar creation and animation on challenging texture cases. Our mesh-based framework successfully models high-fidelity texture details with only 8K vertices, substantially outperforming the Gaussian-based LAM method that requires 80K Gaussian points while still failing to capture fine details.}
    \label{fig:qualitative_res}
    \vspace{-4mm}
\end{figure*}

\section{Experiments}
\label{sec:experiments}

\subsection{Experiments Setting}

\noindent\textbf{Implementation Details.}
Our framework is implemented in PyTorch using a frozen DINOv2 backbone. The transformer has 2 layers with 16 attention heads and $C_t=1024$ hidden dimensions. 
Training runs for 100 epochs with Adam, cosine annealing, and linear warm-up. Hyperparameters are set empirically: learning rate $2\times10^{-4}$. Semantic deformation constraints apply clipping ranges as: $\delta_{\text{hair}}=0.08$, $\delta_\text{neck}=0.02$, and $\delta_{\text{face}}=0.003$. See the Supplementary Material for more implementation details.

\noindent\textbf{Datasets and Preprocessing.}
We train our model on the VFHQ dataset~\cite{vfhq}, which contains 15,204 video clips (3M frames) from diverse interview scenarios. Following established practices~\cite{GAGAvatar,GaussianAvatar}, we process each frame by detecting face regions, expanding bounding boxes to capture head and hair context, and resizing to $512\times512$ pixels. To ensure geometric consistency, we track camera poses and FLAME parameters across sequences using the pipeline from~\cite{GaussianAvatar}. Additionally, we perform background removal following~\cite{GPAvatar} and head parsing with~\cite{dinu2022faceparsing} to isolate subjects and minimize background interference during training. For evaluation, we use the official VFHQ test split, where the first frame serves as the source image and subsequent frames provide driving signals and target views for reenactment evaluation. 

\noindent\textbf{Evaluation Metrics.}
We employ three paired-image evaluation metrics to assess the quality of each animated and rendered image: Peak Signal-to-Noise Ratio (PSNR), Structural Similarity Index (SSIM), and Learned Perceptual Image Patch Similarity (LPIPS). Furthermore, we evaluate the similarity between the rendered images and the real images with the Fréchet Inception Distance (FID)~\cite{gans_fid_2017}, which compares the statistical distributions of features extracted from these images. Identity preservation is measured through cosine similarity of face recognition features (CSIM) on the first frame of each person using~\cite{ArcFace} and the implementation from InsightFace\footnote{https://github.com/deepinsight/insightface}; and motion accuracy is quantified via Average Keypoint Distance (AKD) from facial landmarks~\cite{AKD}. Since our focus is head reconstruction, all metrics are computed within a head region mask obtained using a pretrained head parsing network~\cite{dinu2022faceparsing}. This avoids bias from background variation and ensures fair evaluation of facial and hair appearance.

\subsection{Main Results}

\noindent\textbf{Qualitative Results.} In Fig.~\ref{fig:qualitative_res}, we present comprehensive qualitative comparisons that demonstrate the superior reconstruction capability of our method. Notably, our approach successfully synthesizes high-fidelity details such as tattoos and text with about 8K vertices that are blurry or corrupt in LAM's reconstruction with 80K Gaussian points. This capability stems from our explicit texture map representation and iterative texture projection and unwrapping mechanism. The texture map efficiently stores high-frequency appearance information that Gaussian-based methods struggle to capture without excessive primitive counts. The mesh-based representation naturally preserves topological integrity, while our iterative refinement process ensures stable deformation. These visual results consistently validate that our method achieves substantially higher reconstruction fidelity and better preserves identity-specific characteristics compared to state-of-the-art alternatives.

\begin{table}
    \centering
    \caption{Quantitative evaluation on one-shot 3D head avatar creation, where w/ denotes with and w/o denotes without.}
    \setlength{\tabcolsep}{2.pt}
    \label{tab:Quant_sampled_testset}
    \scriptsize
    \begin{tabular}{l|l|cccccc}
        \toprule
        3D Rep. &Method & PSNR$\uparrow$ & SSIM$\uparrow$ & LPIPS$\downarrow$ & AKD$\downarrow$ & CSIM$\uparrow$ & FID$\downarrow$ \\
        \midrule
        \multirow{3}{*}{Mesh} & ROME w/ UNet   & 22.850  & 0.874  & 0.098   & 4.98   & 0.681 & 42.542 \\
                              & Ours w/o UNet  & 23.180  & 0.859  & 0.073   & 3.58   & 0.935 & 23.688 \\
                              % & Ours w/ UNet    & \textbf{25.724}  & \textbf{0.884}  & \textbf{0.058}   & \textbf{3.14}   & \textbf{0.948} & \textbf{21.060} \\
                              & Ours w/ UNet    & \textbf{25.233}  & \textbf{0.879}  & \textbf{0.061}   & \textbf{3.24}   & \textbf{0.948} & \textbf{22.699} \\

        \midrule
        \multirow{2}{*}{Gaussian} & LAM+FLAME & 25.082 & 0.879  & 0.077  & 2.07  & 0.879 & 24.270 \\
                                  & LAM+Ours &\textbf{25.889}  & \textbf{0.893}  & \textbf{0.050} &\textbf{2.02}  & \textbf{0.898} & \textbf{22.576} \\
        \bottomrule
    \end{tabular}
    \vspace{-6mm}
\end{table}

\noindent\textbf{Quantitative Results.}
As shown in Table~\ref{tab:Quant_sampled_testset}, our method achieves state-of-the-art performance across multiple evaluation metrics, demonstrating significant advantages in both reconstruction quality and identity preservation. In the mesh-based comparison, our approach with UNet rendering achieves the best results, with PSNR of 25.233, SSIM of 0.879, and LPIPS of 0.061, substantially outperforming ROME~\cite{rome2022}, another mesh-based method and our variant without UNet. Remarkably, even without the neural renderer, our method maintains competitive performance while being more efficient. The identity preservation metrics further validate our approach, with CSIM reaching 0.948, indicating excellent fidelity to the source identity. More importantly, when integrated with Gaussian-based reconstruction, our mesh initialization (LAM+Ours) achieves the overall best performance across all metrics, including PSNR of 25.889, SSIM of 0.893, and LPIPS of 0.050, demonstrating that our reconstructed mesh serves as a superior geometric prior for downstream applications. These results consistently show that our method not only excels in mesh-based reconstruction but also provides a foundation for enhancing other representation paradigms.

\subsection{More Applications.}

\begin{figure}
\centering
  \includegraphics[width=1\columnwidth, trim={0cm 0cm 0cm 0cm}, clip]{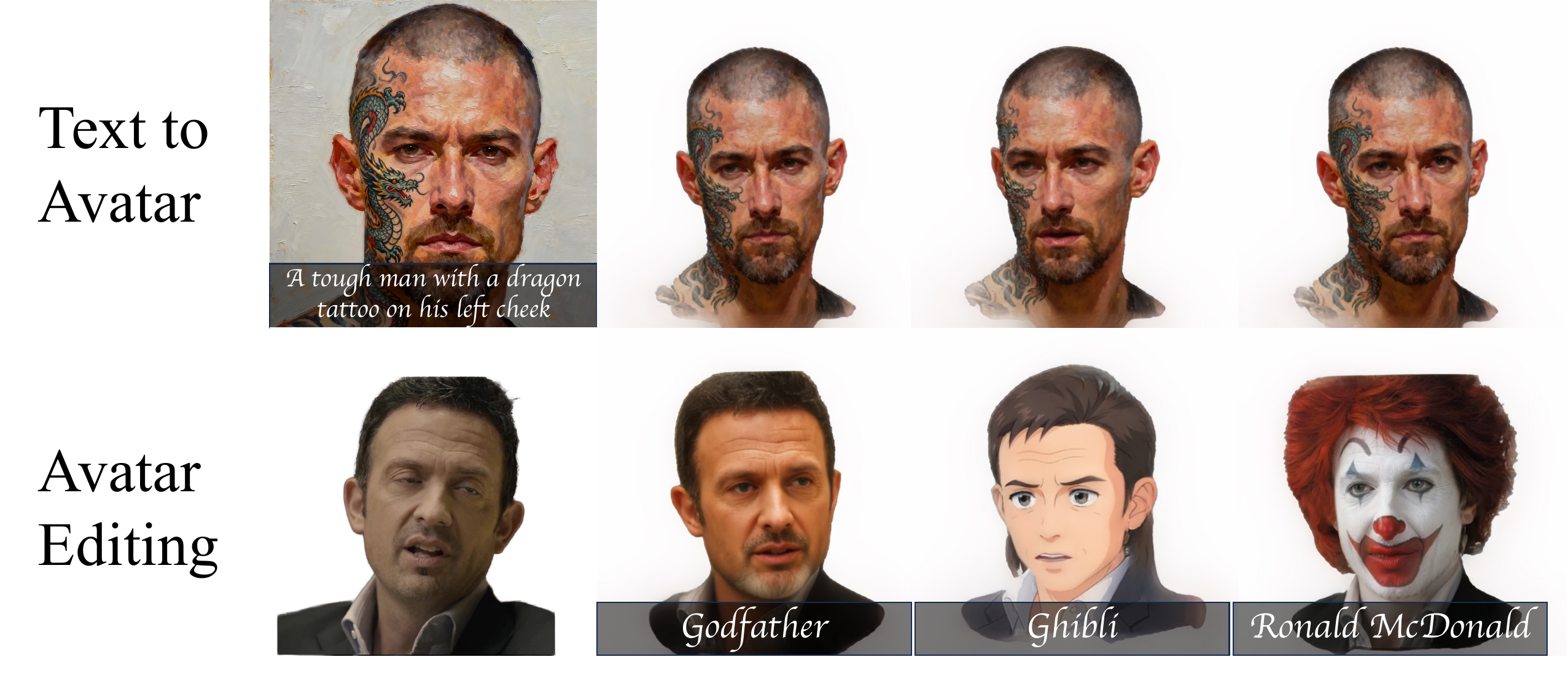}
  % \vspace{2in}
\vspace{-3mm}
\caption{The cross-domain generalization capability of our framework enables easy adaptation to text-to-3D avatar generation and editing, by incorporating a pretrained image generation framework.}
\label{fig:application_text2image}
\vspace{-4mm}
\end{figure}

\noindent\textbf{Text to 3D Avatar Generation.}
As shown in Fig.~\ref{fig:application_text2image}, our framework can be naturally extended to text-conditioned 3D avatar generation by leveraging pretrained text-to-image models~\cite{xie2024sana,wu2025qwenimg,liu2025step1ximg}. Specifically, we first generate a frontal head image using a diffusion model conditioned on text prompts, then feed this synthesized image directly into our network to produce a complete 3D avatar. Remarkably, this two-stage approach demonstrates strong generalization capability, producing coherent 3D shapes and textures that faithfully reflect diverse textual descriptions while maintaining anatomical plausibility. The structured nature of our mesh representation ensures that generated avatars are immediately animatable and topologically sound, significantly outperforming Gaussian-based alternatives that often hard to capture detailed texture with a single feed-forward framework.

\begin{figure}
\centering
  \includegraphics[width=1\columnwidth, trim={0cm 0cm 0cm 0cm}, clip]{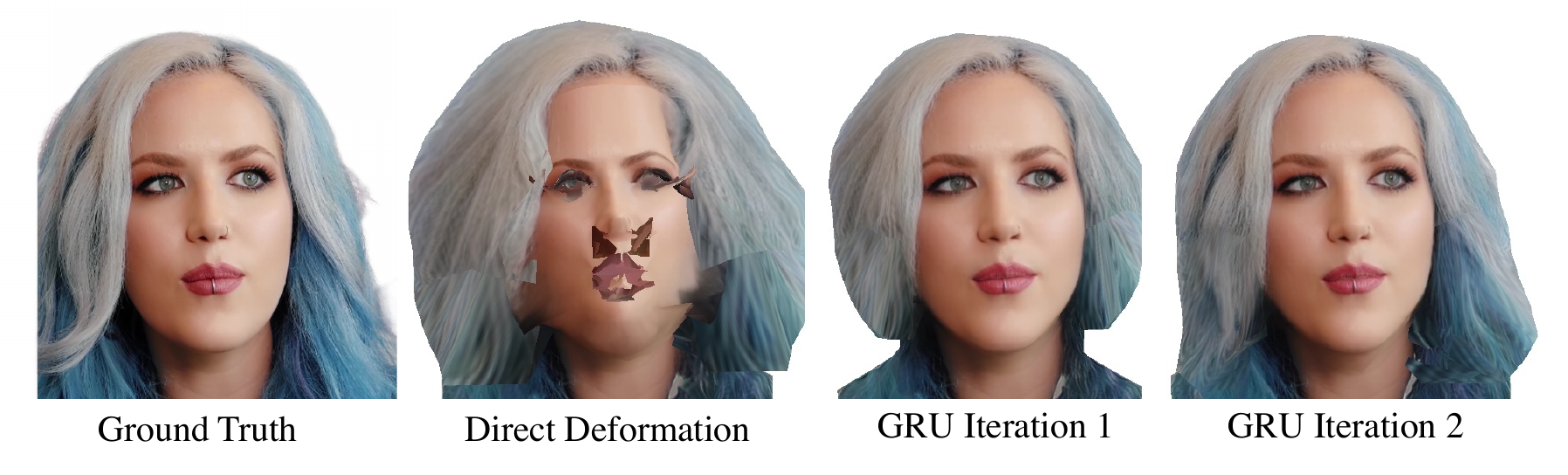}
\caption{Effect of iterative mesh decoding. Without our GRU-based progressive deformation, direct regression suffers from severe mesh collapse and topological corruption. Our approach enables stable coarse-to-fine deformation that preserves mesh integrity while recovering better geometry.}
\label{fig:ablation_gruiter}
\vspace{-2mm}
\end{figure}

\noindent\textbf{Efficient Avatar Style Transfer and Editing.}
Unlike previous 3D editing frameworks~\cite{FreditorECCV,instructnerf2nerf,gaussianeditor} that require training for each editing style, our framework enables efficient avatar style transfer by combining pretrained image style editing models with our reconstruction pipeline in a single forward pass. We first apply style transfer and editing algorithms~\cite{xie2024sana,wu2025qwenimg,liu2025step1ximg} to modify the input image's appearance, then process the stylized image through our network to generate a corresponding 3D avatar. This approach leverages the powerful editing capabilities of 2D style transfer methods while benefiting from the 3D consistency of our mesh representation. The resulting avatars maintain the desired artistic style across all viewpoints and remain fully animatable, demonstrating the practical advantages of our representation for creative applications and content personalization.

\subsection{Ablation Studies}

\begin{figure}
\centering
\includegraphics[width=1.0\columnwidth, trim={0cm 0cm 0cm 0cm}, clip]{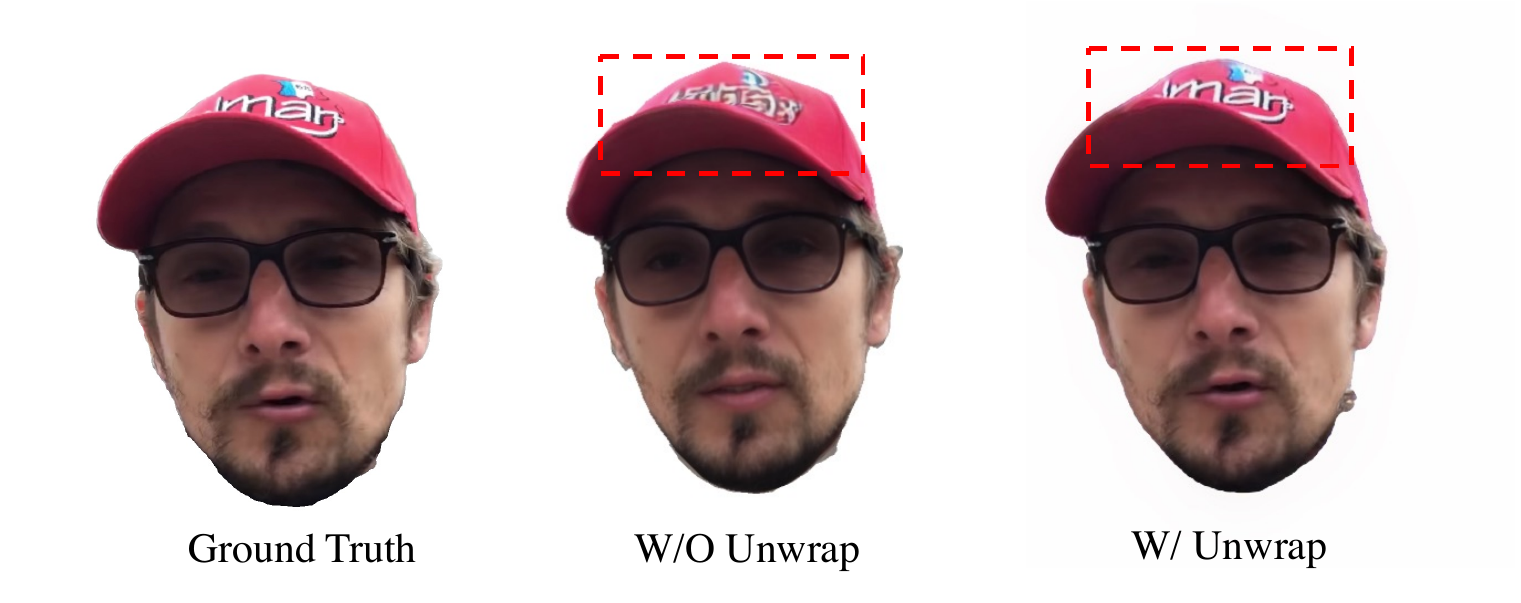}
\vspace{-3mm}
\caption{Effect of texture space reprojection. Without our reprojection mechanism (middle), texture synthesis lacks direct visual guidance, resulting in a blurry appearance, while ours enable high-fidelity texture details.}
\label{fig:ablation_unwrap}
\end{figure}

\begin{figure}
\centering
\includegraphics[width=1.\columnwidth, trim={0cm 0cm 0cm 0cm}, clip]{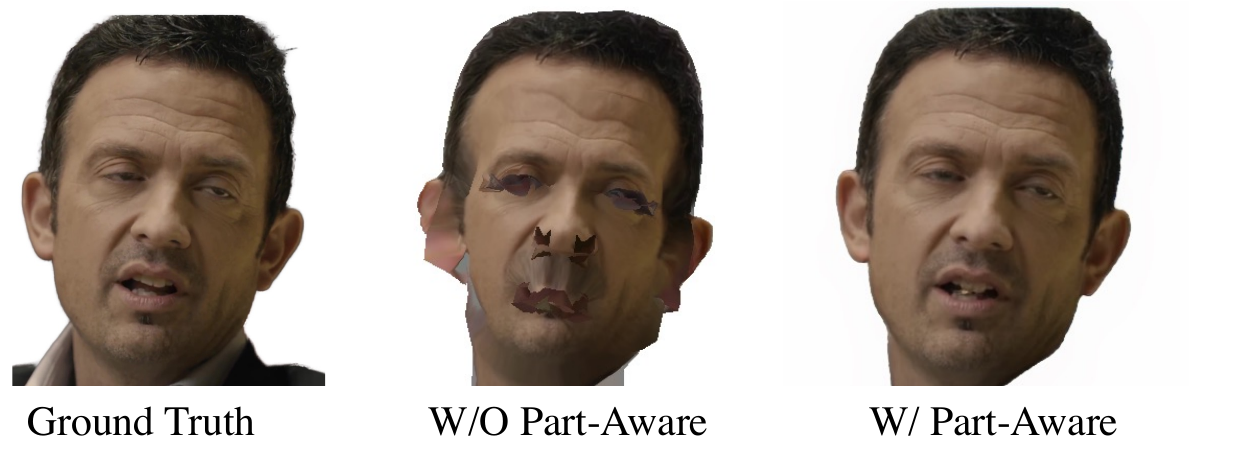}
\vspace{-3mm}
\caption{Effect of part-aware deformation. Without semantic constraints, deformation leads to implausible facial structures. Our part-aware approach preserves anatomical correctness while allowing flexible deformation of hair and accessories.}
\label{fig:ablation_part_deform}
\vspace{-3mm}
\end{figure}

\noindent\textbf{Effect of dual mesh vertex and texture map representation.} 
Table~\ref{tab:quant_ablation} demonstrates the critical importance of our dual representation approach. 
 \begin{wraptable}{r}{0.56\linewidth}
    \vspace{-2.5mm}
    \caption{Quantitative ablation study of different design choices.}
    \centering
    \setlength{\tabcolsep}{2.pt}
    \label{tab:quant_ablation}
    \scriptsize
    \begin{tabular}{l|ccc}
        \toprule
        Method & PSNR$\uparrow$ &LPIPS$\downarrow$ & FID$\downarrow$ \\
        \midrule
        Ours-Full           & 25.23  & 0.061  & 22.699 \\
        w/o Texture Map     & 18.09  & 0.126  & 74.083  \\
        w/o GRU             & 23.08  & 0.081 & 26.397 \\
        w/o Unwrapping      & 22.98  & 0.089  & 29.428 \\
        w/o P.A. Deform.    & 22.72  & 0.096  & 32.405 \\
        w/o UNet            & 23.18  & 0.073  & 23.688 \\
        \midrule
        GRU-1iter           & 23.10 & 0.077 & 25.747 \\
        GRU-2iter           & 25.23 & 0.061 & 22.699  \\
        GRU-3iter           & 25.38 & 0.063 & 23.431 \\
        \bottomrule
    \end{tabular}
    \vspace{-2mm}
\end{wraptable}
When replacing the texture map with per-vertex colors (w/o Texture Map), we observe a dramatic performance drop. This substantial degradation confirms that vertex-attached colors are insufficient for capturing high-frequency appearance details, particularly for complex regions like hair and facial features. The texture map representation proves essential for storing fine-scale appearance information efficiently while maintaining a lightweight mesh structure.

\noindent\textbf{Effect of GRU-based iterative refinement process.}
Our iterative GRU-based refinement mechanism provides substantial performance gains over single-pass regression.
Removing the GRU module (w/o GRU) results in a significant performance drop. As shown in Fig.~\ref{fig:ablation_gruiter}, without GRU regressors, a single pass of shape deformation leads to a collapsed mesh surface. The ablation of iteration numbers further reveals that two iterations achieve the optimal balance between performance and efficiency, with GRU-2iter getting the best results. This demonstrates that progressive refinement is crucial for stable mesh deformation and high-fidelity texture synthesis, while excessive iterations may lead to computational inefficiency.

\noindent\textbf{Effect of texture space projection and unwrapping.}
The texture space projection and unwrapping mechanism play a vital role in grounding our reconstruction in observable visual evidence. Removing this component (w/o Unwrapping) causes notable performance degradation. This demonstrates that direct visual guidance from the input image is essential for realistic texture synthesis and provides valuable geometric constraints. The unwrapping mechanism effectively bridges the gap between 2D observations and 3D reconstruction, ensuring appearance consistency and improving overall reconstruction fidelity, which is also shown in Fig.~\ref{fig:ablation_unwrap}.

\noindent\textbf{Effect of part-aware deformation.}
Our part-aware deformation constraints are crucial for maintaining anatomical plausibility while allowing sufficient flexibility for non-rigid components. Removing these constraints (w/o P.A. Deform.) results in a performance drop, with PSNR decreasing and FID worsening. This substantial degradation confirms that without semantic guidance, the model struggles to balance deformation flexibility with structural integrity, particularly in critical facial regions, which is shown in Fig.~\ref{fig:ablation_part_deform}. The part-aware approach enables targeted deformation control, preserving animation capabilities while capturing complex hair geometry.
\section{Conclusion}
\label{sec:conclusion}

We present a feed-forward framework for one-shot animatable 3D head reconstruction from a single image. Our dual-branch architecture jointly refines mesh geometry and texture with features extracted from a shared transformer backbone, enabling high-fidelity, topology-preserving deformation via iterative GRU-based updates. A reprojection mechanism provides direct visual guidance by unwrapping the input image onto the evolving mesh, ensuring appearance realism without post-processing. Experiments show our method outperforms state-of-the-art in both reconstruction quality and animation fidelity, while running within a second.

\paragraph{Limitations.}
Our method relies on the FLAME for animation. Like other 3D-driven methods, e.g.,~\cite{LAM_Sig25,GAGAvatar}, it is limited by FLAME's expressiveness and cannot generate fine-grained expressions like dynamic wrinkle and tongue movement. Also, the training and animation quality are affected by the accuracy of FLAME parameter estimation algorithms. We believe that developing more expressive 3DMMs (e.g., differentiable MetaHuman) and more accurate 3DMM estimators will resolve this problem.

{
    \small
    \bibliographystyle{ieeenat_fullname}
    \bibliography{main}
}

\clearpage
\appendix
% \clearpage
\setcounter{page}{1}
\setcounter{table}{2}
\setcounter{figure}{6}

\section{More Results}
\label{sec:sup_more_res}
\begin{figure}[t]
    \centering
    \includegraphics[width=.9\linewidth]{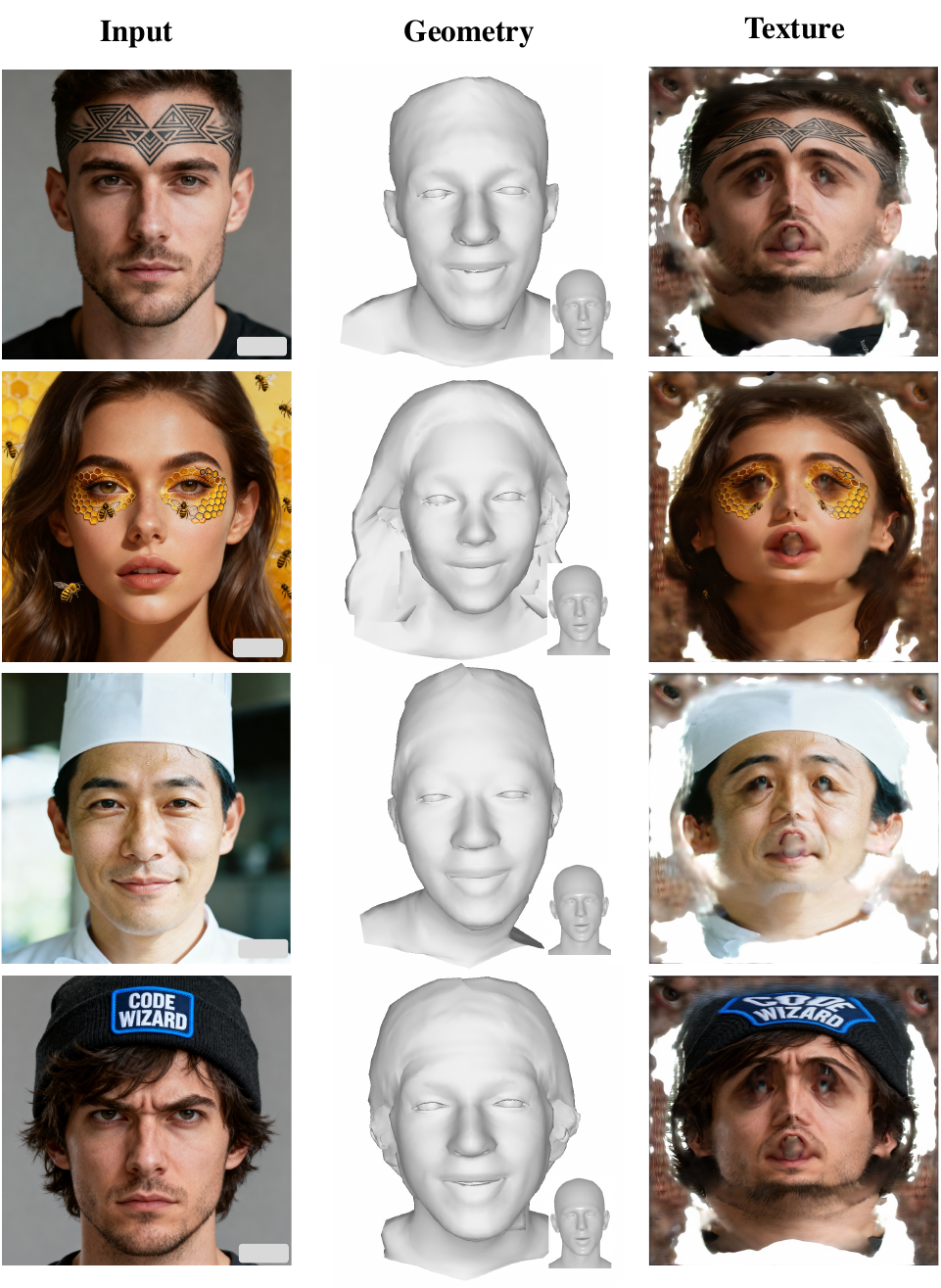}
    \caption{Reconstructed geometry and texture map visualization. Our method produces a mesh with high-fidelity texture, even when animated to an expression different from the input image. The sharp, consistently unwrapped texture details demonstrate the effectiveness of our dual-branch design.}
    \label{fig:geometry_texture}
\end{figure}

\paragraph{Visualization of Reconstructed Geometry and Texture Map.}
As shown in Fig.~\ref{fig:geometry_texture}, our method produces high-quality geometry and texture outputs from a single input image. The reconstructed mesh faithfully captures the subject's facial structure, hair volume, and accessory shapes, demonstrating significant deformation from the FLAME template while maintaining topological integrity. The corresponding texture map displays sharp, high-resolution details with consistent color and lighting, effectively transferring the input appearance to the UV atlas. This visualization underscores the advantage of our explicit mesh and texture representation, which jointly enables detailed geometry reconstruction and photorealistic texture synthesis in a compact and efficient form.

\paragraph{Animation Results Visualization.}
We provide visualization comparisons of the in-the-wild animation results from the proposed framework and previous state-of-the-art methods, LAM~\cite{LAM_Sig25}, in the supplementary videos. We can see from the videos that our mesh-based framework can better preserve identity and texture details, such as texts and tattoos on the head.

\begin{figure}[t]
  \centering
  \includegraphics[width=1.\linewidth]{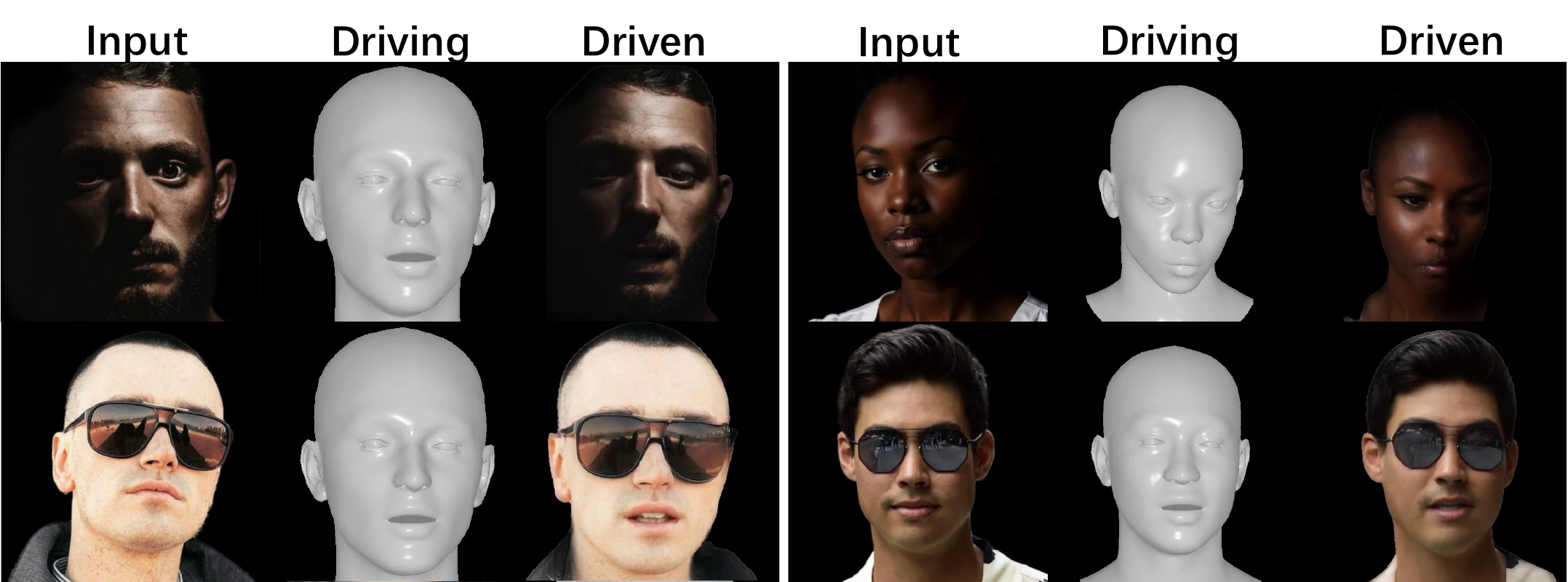}
   \caption{In the wild challenging lighting and occlusion. }
   \label{fig:rebut_low_input}
\end{figure}

\paragraph{Experiments on low-quality, in-the-wild inputs.}
Fig.~\ref{fig:rebut_low_input} shows our method with low-quality out-of-domain (OOD) input. Our method can generalize to both OOD AI-generated and real-world in-the-wild low-quality images.

\paragraph{More baselines.} 

\begin{figure}[t]
  \centering
   \includegraphics[width=1.0\linewidth]{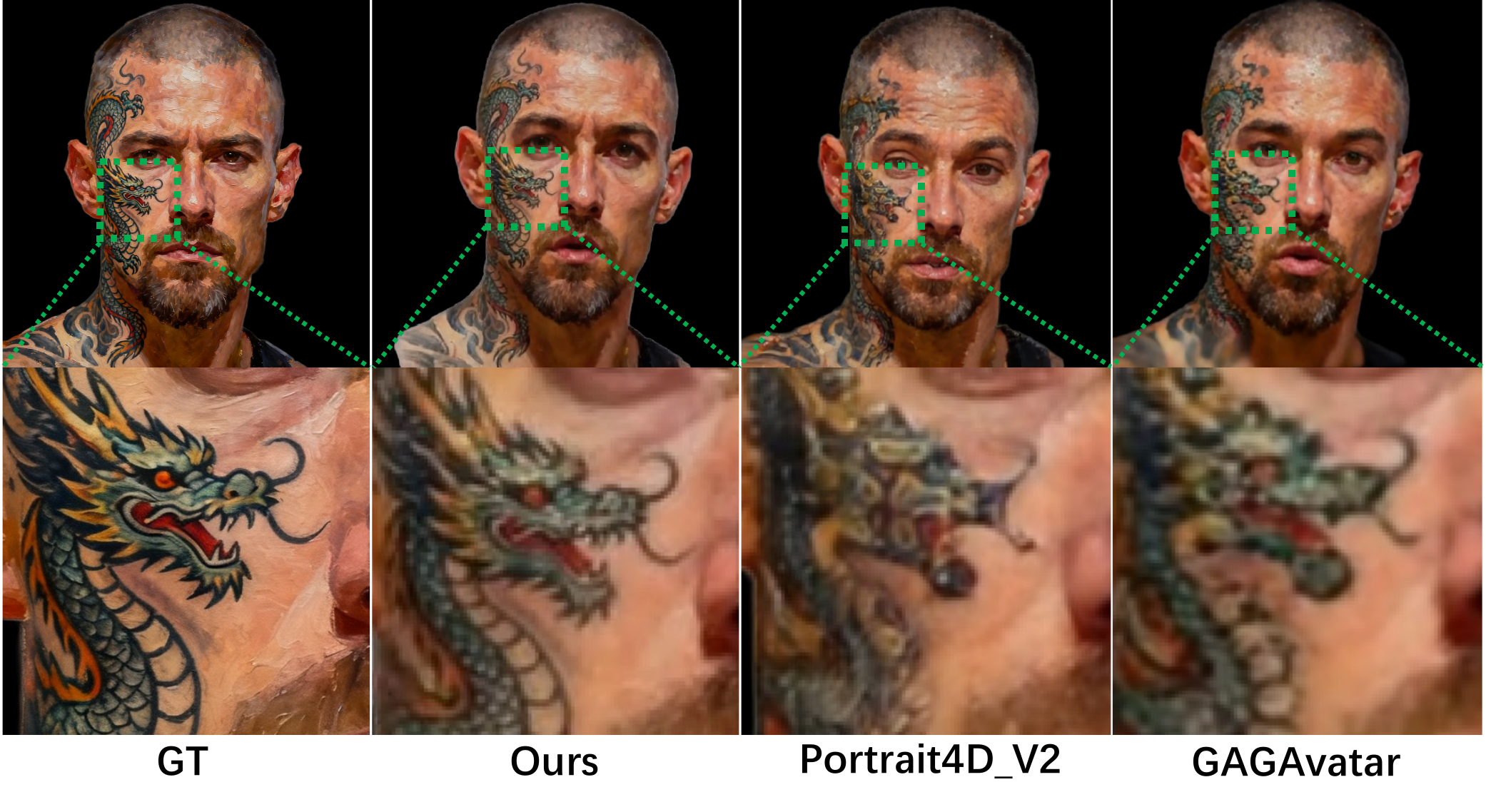}
   \caption{Qualitative comparison with more baselines.}
   \label{fig:rebut_more_cmp}
\end{figure}

\setlength{\columnsep}{5pt}
\begin{wraptable}{r}{4.cm}
    % \vspace{-2.5mm}
    \centering
    \captionsetup{font={footnotesize}}
    \setlength{\abovecaptionskip}{20pt}  
    \vspace{-4.6mm}
    \caption{Quantitative comparison with more baselines.}
    \vspace{-6.8mm}
    \setlength{\tabcolsep}{1.pt}
    \label{tab:rebut_quant_testset}
    \footnotesize
    \resizebox{\linewidth}{!}{\begin{tabular}{l|cccccc}
        \toprule
        Method & PSNR$\uparrow$ & SSIM$\uparrow$ & LPIPS$\downarrow$ & FID$\downarrow$ \\
        \midrule
         Portrait4Dv2   & 24.650  & 0.835  & 0.076  & 30.137 \\
         GAGAvatar      & 25.020  & 0.864  & 0.066  & 25.014 \\
         Ours           & \textbf{25.233}  & \textbf{0.879}  & \textbf{0.061} & \textbf{22.699} \\ 
        \bottomrule
    \end{tabular}}
    \vspace{-6mm}
\end{wraptable}
We present the qualitative comparisons with more baselines in Fig.~\ref{fig:rebut_more_cmp}, including GAGAvatar~\cite{GAGAvatar} and Portrait4Dv2~\cite{Portrait4D_v2_DengWW24}. The quantitative results are shown in Tab.~\ref{tab:rebut_quant_testset}. Compared with previous work, our methods better reconstruct the texture details of the person.

\paragraph{Reconstruction Speed.}
Our method demonstrates significant computational efficiency, reconstructing a complete 3D avatar in only \textit{0.7 seconds} from a single input image. In comparison, the Gaussian-based LAM~\cite{LAM_Sig25} requires 1.4 seconds for 20K Gaussians and 5.8 seconds for 80K Gaussians. This substantial speedup stems from our efficient mesh-based representation: our model operates on only 5K vertices and a compact texture token grid (4K tokens) for cross-attention, whereas Gaussian-based methods must process orders of magnitude more primitives (20K or 80K) to reconstruct details but still fail on high-frequency textures like texts and tattoos. The lightweight nature of our representation reduces both memory footprint and computational overhead, enabling fast avatar generation without sacrificing reconstruction quality.

\section{More Implementation Details}
\label{sec:sup_implement}
Our framework utilizes a learnable token grid $T_0$ to extract texture features from the input image to reconstruct the texture map. We initialize $T_0$ to a shape of 64x64, with $H_t=W_t=64$, and decode them into a $1024\times1024\times3$ texture map, by setting the hyperparameters $H_a = W_a = 1024$. The convolution operation $\varphi$ in the iterative texture synthesis blocks, $GRU_\text{tex}$ denoted in Formula 4, is implemented with 2-layer convolutions with kernel size 3. The MLP operation $\vartheta$, for iterative geometric deformation denoted in Formula 5, is implemented with two-layer MLPs.

\section{Ethical Impact}
Generating realistic 3D avatars from single images raises ethical concerns regarding privacy, consent, and potential misuse for deceptive content. While enabling positive applications in virtual communication, our method also carries risks of unauthorized identity replication and manipulation. We stress the importance of ethical frameworks with consent mechanisms and safeguards to ensure responsible use. Continued dialogue and proactive measures are crucial to mitigate harms while preserving societal benefits.

\end{document}